%% file: main.tex
\documentclass[10pt,twocolumn,letterpaper]{article}

\usepackage{cvpr}              % To produce the CAMERA-READY version
\input{preamble}

\definecolor{cvprblue}{rgb}{0.21,0.49,0.74}
\usepackage[accsupp]{axessibility} % Improves PDF readability for those with disabilities.
\usepackage[pagebackref,breaklinks,colorlinks,citecolor=cvprblue]{hyperref}
\usepackage{url}
\usepackage{times}
\usepackage{epsfig}
\usepackage{graphicx}
\usepackage{amsmath}
\usepackage{amssymb}
\usepackage{multirow}
\usepackage{booktabs}
\usepackage{wrapfig}
\usepackage{caption}
\def\paperID{1862}
\def\confName{CVPR}
\def\confYear{2024}

\title{Token Transformation Matters: \\Towards Faithful Post-hoc Explanation for Vision Transformer}

\author{
    \textbf{Junyi Wu}$^1$, \textbf{Bin Duan}$^1$, \textbf{Weitai Kang}$^1$, \textbf{Hao Tang}$^2$,
    \textbf{Yan Yan}$^{1,}$\footnotemark[2] \\
    $^1$Department of Computer Science, Illinois Institute of Technology, USA \\
    $^2$Robotics Institute, Carnegie Mellon University, USA \\
\tt\small{
\{jwu125, bduan2, wkang11\}@hawk.iit.edu, haotang2@cmu.edu, yyan34@iit.edu
}
}

\begin{document}
\maketitle

\renewcommand{\thefootnote}{\fnsymbol{footnote}}
\footnotetext[2]{Corresponding author}

\input{sec/0_abstract}
\input{sec/1_introduction}
\input{sec/2_relatedwork}

\input{sec/3_analysis}
\input{sec/4_method}

\input{sec/5_experiment}

\input{sec/6_conclusion}

{
    \small
    \bibliographystyle{ieeenat_fullname}
    \bibliography{main}
}

\end{document}

%% file: preamble.tex
\usepackage[dvipsnames]{xcolor}

%% file: sec/0_abstract.tex
\begin{abstract}
While Transformers have rapidly gained popularity in various computer vision applications, post-hoc explanations of their internal mechanisms remain largely unexplored.
% While Vision Transformers, characterized by their growing complexity, excel in various computer vision tasks, the intricacies of their internal dynamics remain largely unexplored.
% To extract visual information, Vision Transformers draw representations from image patches as transformed tokens and then integrate them based on attention weights.
Vision Transformers extract visual information by representing image regions as transformed tokens and integrating them via attention weights.
% However, current explanation methods only focus on attention weights without considering corresponding vectors’ contributions, ignoring the essential information from vector transformations.
% However, current explanation methods only focus on attention weights without considering the corresponding vectors, ignoring the essential information from vector transformations.
However, existing post-hoc explanation methods merely consider these attention weights, neglecting crucial information from the transformed tokens, which fails to accurately illustrate the rationales behind the models' predictions.
To incorporate the influence of token transformation into interpretation, we propose \textbf{TokenTM}, a novel post-hoc explanation method that utilizes our introduced measurement of token transformation effects.
Specifically, we quantify token transformation effects by measuring changes in token lengths and correlations in their directions pre- and post-transformation.
Moreover, we develop initialization and aggregation rules to integrate both attention weights and token transformation effects across all layers, capturing holistic token contributions throughout the model.
% Experimental results demonstrate the superiority of TokenTM compared to state-of-the-art post-hoc explanation methods.
Experimental results on segmentation and perturbation tests demonstrate the superiority of our proposed TokenTM compared to state-of-the-art Vision Transformer explanation methods.
% Vision Transformers have seen a surge in popularity for computer vision tasks. Despite their capabilities, the inherent complexity deems them as black-box models, necessitating interpretability for human understanding.
% However, current attention-based Vision Transformer explanations yield suboptimal results due to an essential issue:
% the attention weights do not fully provide information about vector transformations within the model, making it difficult to faithfully capture the rationales behind predictions.
% This can result in misleading interpretation heatmaps (e.g., confounding foreground objects with background regions).
% To address this, we propose \textbf{VTranM}, a novel explanation method leveraging a \underline{v}ector \underline{tran}sformation \underline{m}easurement.
% The measurement considers changes in vector length and directional correlation to faithfully evaluate transformation effects.
% Furthermore, given the multi-layered structure of Transformers, our approach uses an aggregation framework to integrate attention and vector transformation information across layers, which captures the comprehensive vector contributions over the entire model.
% Experiments on segmentation and perturbation tests demonstrate the superiority of VTranM compared to state-of-the-art explanation methods.
\end{abstract}

%% file: sec/1_introduction.tex
\section{Introduction}
The application of Transformer models has resulted in superior performance in computer vision \cite{vaswani2017attention, carion2020end, dosovitskiy2020image, touvron2021training, he2022masked, liu2022swin}, paving the way for new breakthroughs in various tasks.
However, the inherent complexity of Vision Transformers often renders them black-box models, making understanding the rationales behind their predictions a challenge.
Such a lack of transparency undermines the trustworthiness of decision-making processes \cite{jain2019attention, wiegreffe2019attention, serrano2019attention, deyoung2020eraser, agarwal2022openxai}.
Therefore, it is required to interpret Transformer networks.
% To increase such transparency,
In the visual domain, post-hoc explanation sheds light on the rationale behind a model's predictions using a heatmap over the input pixel space. This approach is both effective and efficient in providing interpretations that are easily understood by humans \cite{mahendran2016visualizing, sundararajan2017axiomatic, zhou2018interpreting, chefer2021transformer}.

There exist two types of post-hoc explanation methods, general traditional explanations \cite{selvaraju2017grad, sundararajan2017axiomatic} and Transformer-specific attention-based explanations \cite{abnar2020quantifying, chefer2021generic}.
Although traditional explanations excel in visualizing important pixels for predictions of MLPs and CNNs, their effectiveness markedly decreases when applied to Vision Transformers, due to the fundamental differences in model architectures.
Therefore, attention-based explanations design new paradigms specific to Transformers, where attention weights play a dominant role \cite{jain2019attention, abnar2020quantifying, chefer2021generic, qiang2022attcat}.
Integrating attention information, these explanation methods offer a promising approach toward Transformer interpretability.

\begin{figure*}[t]
  % \vspace*{-7mm}
  \centering
  \resizebox{0.83\textwidth}{!}{%
  % \centering
  \includegraphics[width=1\textwidth]{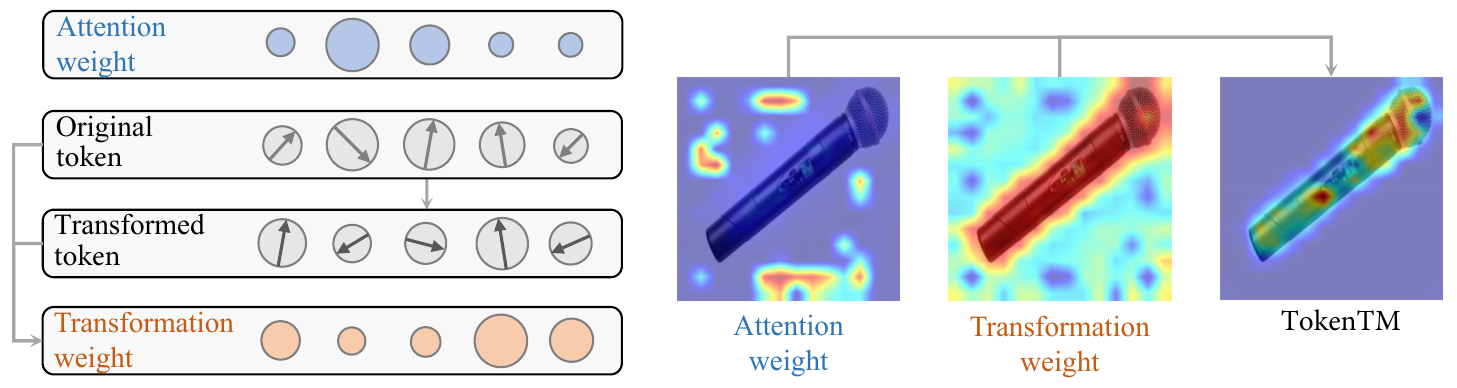}
  }
  \vspace{1mm}
  \caption{Visualization of attention and transformation weights, and the result of our TokenTM that integrates both of them. Circle sizes signify weight magnitudes or token lengths, and arrows indicate directions.
  Transformation weights are derived by our proposed measurement, which evaluates the transformation effects by gauging changes in length and direction.
  Both weights are visualized by heatmaps.
  Solely using attention weights often fails to localize foreground objects and inaccurately highlights noisy backgrounds as rationales.
  In contrast, leveraging additional information from transformation, our method produces object-centric post-hoc interpretations.}
  \vspace{-5mm}
  \label{figure1}
\end{figure*}

Recent advancements in attention-based explanation techniques have outperformed traditional methods on Vision Transformers.
However, the inherent complexity of these models, particularly due to their key components: Multi-Head Self-Attention (MHSA) and Feed Forward Network (FFN) \cite{dosovitskiy2020image}, continues to pose unique challenges in terms of explainability. As elucidated in Section \ref{section3}, we represent the outputs of both MHSA and FFN as weighted sums of transformed tokens, with each token scaled by an attention weight.
Existing attention-based methods simply consider the scaling weights indicated by attention maps as the corresponding tokens' contributions, overlooking the impacts of token transformations.
As shown in Figure \ref{figure1}, attention weights alone misrepresent the contributions from foreground objects or background regions, while transformation information offers a necessary counterbalance.
For instance, even if certain background regions are scaled by high attention weights, their actual contribution can be diminished if they are transformed into smaller or divergent tokens.
Conversely, a foreground object, despite receiving minimal attention weights, can play a pivotal role due to significant transformation within the model.
Given the intertwined dynamics of attention weights and token transformations, there is an imperative need for a comprehensive explanation method that cohesively addresses both factors.

In this paper, we propose \textbf{TokenTM}, a novel post-hoc explanation method for Vision Transformers using token transformation measurement.
We first revisit Vision Transformer layers from a generic perspective, which conceives both MHSA and FFN's outputs as weighted linear combinations of original and transformed tokens.
% paving the way for a comprehensive evaluation of their impacts.
Intuitively, tokens with increased magnitude and aligned orientations will significantly influence the linear combination outcomes.
% as illustrated in/by xxx
Therefore, to quantify the effects of transformations, we introduce a measurement that focuses on two fundamental token attributes: length and direction.
Using the proposed measure, we integrate transformation effects, quantified by transformation weights, with the attention information, ensuring a faithful assessment of token contributions.
Moreover, Vision Transformers consist of sequentially stacked layers, each performing vital token transformations and globally mixing different tokens, a process termed contextualization \cite{vaswani2017attention}.
Recognizing the accumulative nature of these mechanisms, a single-layer analysis remains insufficient \cite{brunner2019identifiability, jain2019attention, abnar2020quantifying}.
To holistically evaluate all layers, our solution encompasses an aggregation framework. Employing rigorous initialization and update rules, our framework yields a comprehensive contribution map, which reveals token contributions over the entire model. Experiments on segmentation and perturbation tests show that our TokenTM outperforms state-of-the-art methods.

In summary, our contributions are as follows:
\textbf{(i)}
We explore post-hoc interpretation for Vision Transformer and identify a primary issue: the lack of comprehensive consideration for token transformations and attention weights, which can result in misleading rationales of the model's prediction.
\textbf{(ii)}
We introduce TokenTM, a novel post-hoc explanation method employing token transformation measurement. This measurement regards changes in length and directional correlation, which faithfully assesses the impact of transformations on token contributions.
\textbf{(iii)}
Our approach establishes an aggregation framework that integrates both attention and token transformation effects across multiple layers, thereby capturing the cumulative nature of Vision Transformers.
\textbf{(iv)}
Using the proposed measurement and framework, TokenTM demonstrates superior performance in comparison to existing state-of-the-art methods.

%% file: sec/2_relatedwork.tex
\section{Related Work}
\textbf{General Traditional Explanations.}
Traditional post-hoc explanation methods mainly fall into two groups: gradient-based and attribution-based.
Examples of gradient-based methods are Gradient*Input \cite{shrikumar2016not}, SmoothGrad \cite{smilkov2017smoothgrad}, Deconvolutional Network \cite{zeiler2014visualizing}, Full Grad \cite{srinivas2019full}, Integrated Gradients \cite{sundararajan2017axiomatic}, and Grad-CAM \cite{selvaraju2017grad}. These methods produce saliency maps using the gradient. On the other hand, attribution-based methods propagate classification scores backward to the input \cite{bach2015pixel, lundberg2017unified, shrikumar2017learning, iwana2019explaining, gu2019understanding, gur2021visualization}, based on Deep Taylor Decomposition \cite{montavon2017explaining}.
There are other approaches beyond these two types, such as saliency-based \cite{zhou2016learning, mahendran2016visualizing, dabkowski2017real}, Shapley additive explanation (SHAP) \cite{lundberg2017unified}, and perturbation-based methods \cite{fong2017interpretable}. Although initially designed for MLPs and CNNs, some traditional methods have been adapted for Transformers in recent works \cite{chefer2021transformer, ali2022xai}.
However, without attention weights, these methods still yield suboptimal performance on Vision Transformers.

\noindent\textbf{Transformer-specific Attention-based Explanations.}
A growing line of work in interpretability develops new paradigms specifically for Transformers. Attention maps are widely used in this direction, as they are intrinsic distributions of scaling weights over tokens. Representative methods include Raw Attention \cite{wiegreffe2019attention}, which regards attention as an explanation, Rollout \cite{abnar2020quantifying}, which linearly accumulates attention maps, GAE \cite{chefer2021generic}, a framework for variants of Transformers, ATTCAT \cite{qiang2022attcat}, which formulates Attentive Class Activation Tokens to estimate their importance, and IIA \cite{barkan2023visual} and DIX \cite{barkan2023deep}, which propose path integration using attention.
However, these approaches overlook the relative effects of transformations and fail to faithfully aggregate token contributions across all modules, hindering reliable post-hoc interpretations for Vision Transformers.

%% file: sec/3_analysis.tex
\section{Analysis}\label{section3}
Vision Transformer \cite{dosovitskiy2020image} sequentially stacks $n_L$ layers, each containing an MHSA or an FFN. We first reinterpret these layers generically and then analyze the problem in Vision Transformer explanations.

\subsection{Revisiting Transformer Layers}
From a general viewpoint, MHSA and FFN both process weighted linear combinations of original and transformed tokens. Starting with a reinterpretation of MHSA, we illustrate this shared principle, then we show how FFN emerges as a particular case within the unified formulation.

In MHSA, every token of the input sequence attends to all others by projecting the embeddings to a query, key, and value.
Formally, let $\mathbf{E} \in \mathbb{R}^{n \times d}$ be the embeddings matrix (a sequence of tokens), where $n$ is the number of tokens, and $d$ is the dimensionality of embedding space.
The projections can be expressed as:
\begin{equation}
\mathbf{Q} = \mathbf{E}\mathbf{W^Q}, \quad
\mathbf{K} = \mathbf{E}\mathbf{W^K}, \quad
\mathbf{V} = \mathbf{E}\mathbf{W^V},
\end{equation}
where $\mathbf{W^Q} \in \mathbb{R}^{d \times d_Q}$, $\mathbf{W^K} \in \mathbb{R}^{d \times d_K}$, and $\mathbf{W^V} \in \mathbb{R}^{d \times d_V}$ are parameter matrices.
Subsequently, the attention map $\mathbf{A}$ is computed by:
\begin{equation}
\mathbf{A} = \text{Softmax} \left(\frac{\mathbf{Q} \mathbf{K}^T}{\sqrt{d_Q}}\right) \in \mathbb{R}^{n \times n}.
\end{equation}
Then, contextualization is performed on value $\mathbf{V}$ using attention map $\mathbf{A}$, and another linear transformation further projects the resulting embeddings back to the space $\mathbb{R}^d$:
\begin{equation}
\mathbf{(AV)W^H} \in \mathbb{R}^{n \times d},
\end{equation}
where $\mathbf{W^H} \in \mathbb{R}^{d_V \times d}$.
To obtain the final output, MHSA integrates the results from multiple heads and incorporates them into original embeddings $\mathbf{E}$ from the previous layer by skip-connection \cite{he2016deep}.
The output of MHSA is given by:
\begin{equation}
\text{MHSA}(\mathbf{E}) = \mathbf{E} + \sum_{h=1}^{n_H} \mathbf{(AV)W^H},
\end{equation}
where $n_H$ is the number of heads. We omit the subscript $h$ for simplicity.
Given the associative property of matrix multiplication, we can regard the combination of value projection and multi-head integration as a simple yet equivalent token transformation featured by $\mathbf{\widetilde{W}} = \mathbf{W^{V} W^{H}}$:
\begin{equation}
\mathbf{(AV)W^H} = \mathbf{A E (W^{V} W^{H})} = \mathbf{A (E \widetilde{W})}.
\end{equation}
We then reformulate the MHSA using the definition of transformed embeddings $\mathbf{\widetilde{E}} = \mathbf{E \widetilde{W}}$:
\begin{equation}
\text{MHSA}(\mathbf{E}) = \mathbf{E} + \sum_{h=1}^{n_H} \mathbf{A \widetilde{E}}.
\label{eq6}
\end{equation}
From a vector-level view, each token of the MHSA's output is a weighted sum of original and transformed tokens.
With this insight, the FFN embodies a basic form of `contextualization', where the number of heads $n_H{=}1$ and the attention map $\mathbf{A}{=}\mathbf{I}$ (identity matrix), as contrasted with Eq. \eqref{eq6}.
Utilizing the concept of transformed embeddings $\mathbf{\widetilde{E}}$, the FFN can be expressed as:
\begin{equation}
\text{FFN}(\mathbf{E}) = \mathbf{E} + \mathbf{I\widetilde{E}},
\label{eq7}
\end{equation}
where $\mathbf{\widetilde{E}}$ consists of tokens that have been transformed within the FFN.

\subsection{Problem in Explaining Vision Transformers}
As depicted in Eq. \eqref{eq6} and Eq. \eqref{eq7}, Vision Transformer layers are generically expressed as weighted sums of original and transformed tokens, where each token is multiplied by a scaling weight.
Existing attention-based explanations typically consider these scaling weights as the corresponding tokens' contributions, which overlooks the influences imparted by the tokens themselves.
As illustrated in Figure \ref{figure1}, relying solely on attention weights can misrepresent the contributions from various image patches.
This issue necessitates a comprehensive method to faithfully interpret the inner mechanism of Transformer layers and capture the true rationales behind the predictions.

\begin{figure*}[t]
  \centering
  \includegraphics[width=0.83\textwidth]{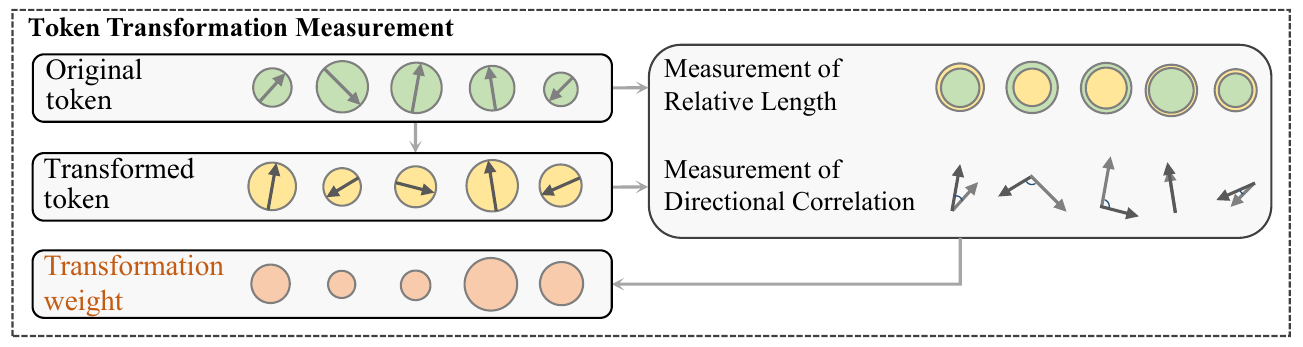}
  \caption{Illustration of our token transformation measurement. We depict original and transformed tokens with circles and arrows. Circle sizes reflect lengths, and arrows denote directions. The effects of token transformation are reflected by the changes in length and direction. Our method considers both properties to evaluate these effects, resulting in the corresponding transformation weights.}
  \vspace{-4mm}
  \label{figure2}
\end{figure*}

%% file: sec/4_method.tex
\section{The Proposed Method}\label{section4}
In this section, we propose TokenTM. We first introduce the background of attention-based explanations for Vision Transformers.
Then, we develop a token transformation measurement, which evaluates the effects of transformed tokens.
Finally, we design an aggregation framework to accumulate both attention and transformation information across all layers.

\subsection{Attention-based Explanations}
Attention-based explanation methods \cite{abnar2020quantifying, chefer2021transformer, chefer2021generic} measure the contribution of each token using the attention information, as expressed by:
\begin{equation}
\mathbf{C} = \mathbf{O} + \mathbb{E}_{h}\left[(\mathbf{\nabla_{\mathbf{A}}}p(c))^+ \odot \mathbf{T}\right],
% \text{ with } 
\mathbf{O} = \mathbf{I}, \mathbf{T} = \mathbf{A}.
\label{eq8}
\end{equation}
% where $\mathbf{O} {=} \mathbf{I}, \mathbf{T} {=} \mathbf{A}$.
Here, $\odot$ is the Hadamard product.
$\mathbf{C} \in \mathbb{R}^{n \times n}$ denotes the contribution map, where $C_{ij}$ represents the influence of the $j$-th input token on the $i$-th output token.
Matrices $\mathbf{O} \in \mathbb{R}^{n \times n}$ and $\mathbf{T} \in \mathbb{R}^{n \times n}$ reflect the contributions from the original and transformed tokens, respectively.
Previous explanation methods quantify $\mathbf{O}$ and $\mathbf{T}$ simply using scaling weights.
Specifically, $\mathbf{I}$ is an identity matrix representing the self-contributions of original tokens, and $\mathbf{A}$ is the attention weights for scaling transformed tokens.
Moreover, $\nabla_{\mathbf{A}}p(c) {=} \frac{\partial p(c)}{\partial \mathbf{A}}$ is the partial derivative of attention map \emph{w.r.t.} the predicted probability for class $c$.
$\mathbb{E}_{h}$ is the mean over multiple heads. Note that it averages across heads using the gradient $\nabla_{\mathbf{A}}p(c)$ to become class-specific, and it removes the negative contributions before averaging to avoid distortion \cite{voita2019analyzing, barkan2021grad, chefer2021transformer, chefer2021generic}.
Previous methods' limitation lies in their assumption of equal influences from the skip connection and the attention weights, neglecting the difference between original and transformed tokens.
% As illustrated in Figure \ref{figure1}, such limitation can miss critical insights from vector transformations.

\subsection{Token Transformation Measurement}
To account for the contributions from transformed tokens, we introduce a measurement that gauges the influence of token transformations and subsequently derives transformation weights.
Using these weights, we recalibrate the matrices $\mathbf{O}$ and $\mathbf{T}$ to encapsulate both attention and transformation insights.
Since MHSA and FFN are depicted in a unified form (see Section \ref{section3}), we can first derive our measurement based on MHSA, and then adapt it for FFN from a generic perspective.

Drawing from foundational principles of token representations, we emphasize two core attributes: length and direction \cite{strang2006linear, kobayashi2020attention}.
The intuition is that tokens with greater lengths and consistent spatial orientations predominantly determine the results of linear combinations.
Thus, our measurement will involve two components, corresponding to these two attributes.
The first component is a length function L$(\mathbf{x}){:}\mathbb{R}^d {\rightarrow} \mathbb{R}^{+}$, which measures the length of a token, whether the original or transformed.
Mathematically, we instantiate L using $L^2$ norm of the embedding space $\mathbb{R}^d$, \emph{i.e.,}
$\text{L}(\mathbf{x}) = \Vert \mathbf{x} \Vert_2$.
% \begin{equation}
% \text{L}(\mathbf{x}) = \Vert \mathbf{x} \Vert_2 = \left(\sum_{i=1}^{d} \vert x_i \vert^2\right)^{\frac{1}{2}},
% \end{equation}
% where $x_i$ is the $i$-th element of vector $\mathbf{x}$.
Considering both the length measurement L and the attention weights, we reintroduce $\mathbf{O}$ and $\mathbf{T}$ as:
\begin{equation}
\mathbf{O}
= \mathbf{I} \cdot \text{diag}(\text{L}(\mathbf{E}_1), \text{L}(\mathbf{E}_2), \ldots, \text{L}(\mathbf{E}_n)),
\end{equation}
\begin{equation}
\mathbf{T}
= \mathbf{A} \cdot \text{diag}(\text{L}(\mathbf{\widetilde{E}}_1), \text{L}(\mathbf{\widetilde{E}}_2), \ldots, \text{L}(\mathbf{\widetilde{E}}_n)).
\end{equation}
Note that $C_{ij}$ indicates the contribution of the $j$-th input token. Thus we apply the function L column-wise.
This approach accounts for both the attention weights and the tokens themselves.
To evaluate the changes in contributions after transformations, we now regard the original tokens as reference units and analyze the relative effects of transformations.
This is done by normalizing $\mathbf{O}$ and $\mathbf{T}$ column-wise \textit{w.r.t.} the lengths of the original tokens. As a result, we obtain:
\begin{equation}
\mathbf{O} = \mathbf{I}, \quad
\mathbf{T}
= \mathbf{A} \cdot \text{diag}\left(\frac{\text{L}(\mathbf{\widetilde{E}}_1)}{\text{L}(\mathbf{E}_1)}, \frac{\text{L}(\mathbf{\widetilde{E}}_2)}{\text{L}(\mathbf{E}_2)}, \ldots, \frac{\text{L}(\mathbf{\widetilde{E}}_n)}{\text{L}(\mathbf{E}_n)}\right).
\end{equation}
In this formulation, $\mathbf{O}$ uses the identity matrix to represent the contributions from original tokens as basic reference units.
Meanwhile, $\mathbf{T}$ discerns the relative influences of transformed tokens compared to the original ones using the ratio of their lengths.
As detailed in Section \ref{subsection4.3}, this approach allows us to initialize a contribution map based on the lengths of the model's input tokens, and then iteratively update the map across layers.
The update employs the ratios of token effects between consecutive layers, tracing the evolution of transformations within the model.
This framework ensures that our analysis remains grounded to the initial input to the model, yet dynamically adapts to every token transformation and contextualization encountered during the inference process.

\begin{figure*}[t]
  \centering
  \includegraphics[width=0.85\textwidth]{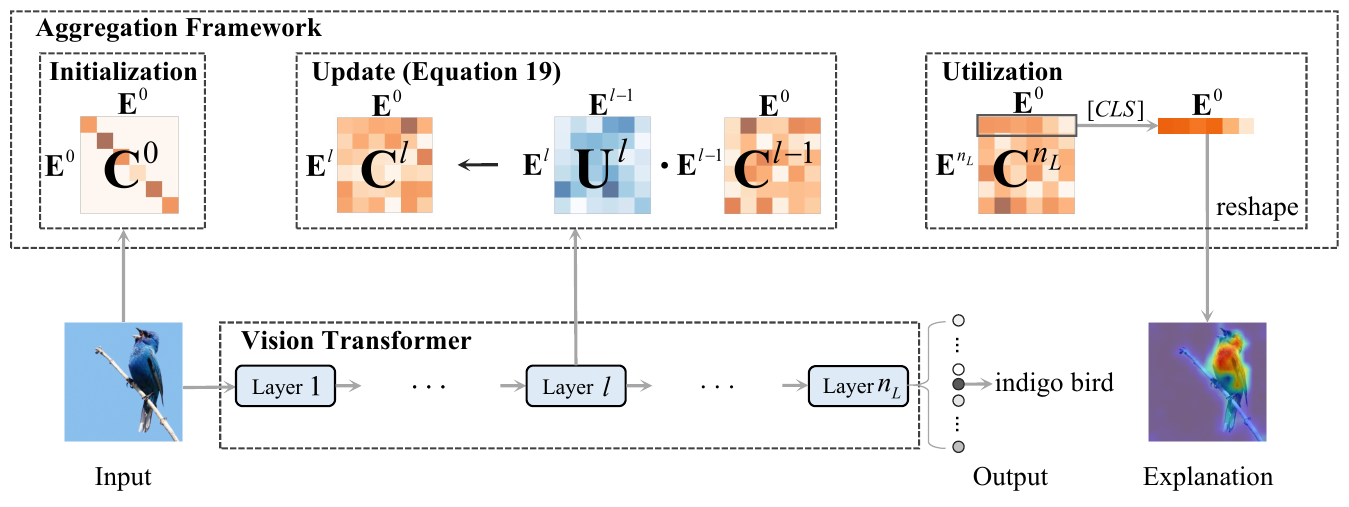}
  \caption{Illustration of our aggregation framework and the explanation pipeline. The overall contribution map is initialized by input token lengths and is updated using our $\mathbf{U}^l$ to trace token evolution across layers.
  In $\mathbf{C}^{l}$, each $i$-th row represents the influences of input tokens $\mathbf{E}^0$ on the output of the $l$-th layer $\mathbf{E}^l$.
  For $\mathbf{C}^{n_L}$, the row \textit{w.r.t.} $[CLS]$ token is extracted and reshaped to produce the final explanation map.}
  \vspace{-6mm}
  \label{figure3}
\end{figure*}

Our second component focuses on directions.
Beyond adjusting lengths, token transformations also influence directions \cite{ethayarajh2019contextual}.
A transformed token that stays directionally closer in representation space is expected to have a stronger correlation with its original counterpart.
To quantify this directional correlation, we introduce a function C$(\mathbf{x}, \mathbf{\widetilde{x}}){:}\mathbb{R}^d {\times} \mathbb{R}^d {\rightarrow} \mathbb{R}$.
Mathematically, we employ Cosine similarity, which measures the angle between a pair of tokens, \emph{i.e.,}
$\text{C}(\mathbf{x}, \mathbf{\widetilde{x}}) = \text{cos}\langle\mathbf{x}, \mathbf{\widetilde{x}}\rangle.$
Next, we will use the function C to complement the length factors in matrix $\mathbf{T}$.
Simply multiplying C as a coefficient may introduce negative values to the contribution map, which will distort the signs of contributions through aggregation \cite{springenberg2014striving, selvaraju2017grad, barkan2021grad, chefer2021transformer, chefer2021generic}.
Inspired by \cite{liu2021adaptive}, we propose the Normalized Exponential Cosine Correlation (NECC).
This measurement is normalized to emphasize the relative magnitudes of each correlation instead of the polarity, thereby serving as an effective positive weighting factor.
For a sequence of original tokens $\mathbf{S_E} = (\mathbf{E}_1, \mathbf{E}_2, \ldots, \mathbf{E}_n)$ and their transformed counterparts $\mathbf{S_{\widetilde{E}}} = (\mathbf{\widetilde{E}}_1, \mathbf{\widetilde{E}}_2, \ldots, \mathbf{\widetilde{E}}_n)$, the weighting factor for the $i$-th pair is given by:
\begin{equation}
\text{NECC}(i)
= \frac{\text{exp}(\text{C}(\mathbf{E}_i, \mathbf{\widetilde{E}}_i))}{\sum_{k=1}^{n} \text{exp}(\text{C}(\mathbf{E}_k, \mathbf{\widetilde{E}}_k))}.
\end{equation}
Treating the original tokens as reference units, NECC is proportional to the extent to which each transformed token correlates with its corresponding original token.
Finally, employing two components of our transformation measurement, we define the transformation weights $\mathbf{W}$:
\begin{equation}
\mathbf{W} = 
\text{diag}\left(\frac{\text{L}(\mathbf{\widetilde{E}}_1)}{\text{L}(\mathbf{E}_1)} \text{NECC}(1), \ldots, \frac{\text{L}(\mathbf{\widetilde{E}}_n)}{\text{L}(\mathbf{E}_n)} \text{NECC}(n)\right),
\end{equation}
and the update map $\mathbf{U}$ for MHSA, incorporating both attention and transformation information:
\begin{equation}
\mathbf{U} = \mathbf{O} + \mathbb{E}_{h}\left[(\mathbf{\nabla_{\mathbf{A}}}p(c))^+ \odot \mathbf{T}\right],
\label{eq14}
\end{equation}
with
\begin{equation}
\mathbf{O} = \mathbf{I}, \quad
\mathbf{T}
= \mathbf{A} \cdot \mathbf{W}.
\label{eq15}
\end{equation}
Here, $U_{ij}$ denotes the influence of the $j$-th input token on the $i$-th output token of the MHSA layer.
This formulation balances both length and direction, reflecting how much of the original information is retained or altered in the transformed tokens, to faithfully evaluate their contributions.

We now adapt our established approach to the FFN layer.
Recall that FFN also involves a weighted sum of original and transformed tokens (see Eq. \eqref{eq7}).
For `contextualization' in FFN, there is only a single-head and the weighted combination is performed locally.
This simpler form can be easily reflected by discarding the gradient-weighted multi-head integration from Eq. \eqref{eq14} and changing the attention map $\mathbf{A}$ in Eq. \eqref{eq15} to an identity matrix.
We formulate the update map for the FFN layer:
\begin{equation}
\mathbf{U} = \mathbf{O} + \mathbf{T},
\end{equation}
with
\begin{equation}
\mathbf{O} = \mathbf{I}, \quad
\mathbf{T}
= \mathbf{I} \cdot \mathbf{W}.
\end{equation}
Here, $U_{ij}$ denotes the influence of the $j$-th input token on the $i$-th output token of the FFN layer.
As illustrated by Figure \ref{figure3}, update maps capture the relative effects and will serve as refinements to the overall contribution map as it aggregates across multiple layers.

\subsection{Aggregation Framework}\label{subsection4.3}
In Vision Transformers, the influences of token transformations and contextualizations do not merely reside within isolated layers but accumulate across them.
For a comprehensive contribution map that reflects the roles of initial tokens corresponding to input image regions, it is necessary to assess the relationships between input tokens and their evolved counterparts in deeper layers. To this end, we introduce an aggregation framework that cohesively measures the combined influences of contextualization and transformation across the entire model.

\begin{figure*}[t]
  \centering
% \begin{minipage}{0.156\linewidth}
%   \centering
%   Grad-CAM
% \end{minipage}
% \begin{minipage}{0.156\linewidth}
%   \centering
%   Trans. Attr.
% \end{minipage}
% \begin{minipage}{0.156\linewidth}
%   \centering
%   Raw Att.
% \end{minipage}
% \begin{minipage}{0.156\linewidth}
%   \centering
%   Rollout
% \end{minipage}
% \begin{minipage}{0.156\linewidth}
%   \centering
%   GAE
% \end{minipage}
% \begin{minipage}{0.156\linewidth}
%   \centering
%   \textbf{Ours}
% \end{minipage}
  \includegraphics[width=0.71\linewidth]{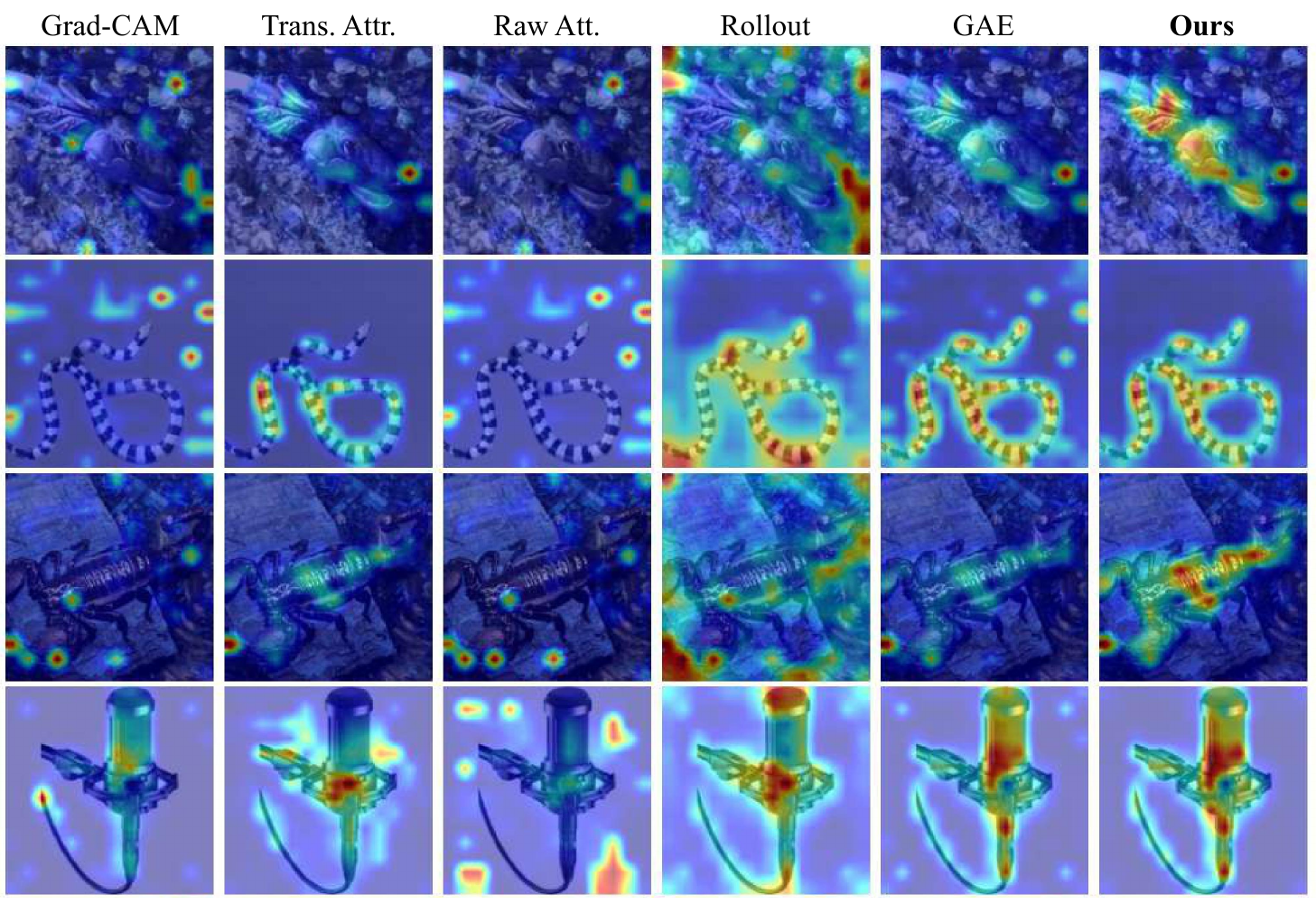}
  \caption{Visualizations of post-hoc explanation heatmaps. Our method captures more object-centric regions as the rationales.}
  \vspace{-0.4cm}
  \label{figure4}
\end{figure*}

\subsubsection{Initialization}
We denote the overall contribution map by $\mathbf{C} \in \mathbb{R}^{n \times n}$, where $n$ is the number of tokens.
The $C_{ij}$ will faithfully accumulate the influence of the $j$-th input token on the $i$-th output token.
At the initial state, before entering Transformer layers, each input token simply contains itself without contextualization or transformation. We represent this state using initial tokens' lengths, as depicted in Figure \ref{figure3}.
Specifically, we initialize the map as a diagonal matrix $\mathbf{C}^0$:
\begin{equation}
\mathbf{C}^0 = 
\text{diag}(\text{L}(\mathbf{E}^0_1), \text{L}(\mathbf{E}^0_2), \ldots, \text{L}(\mathbf{E}^0_n)),
\end{equation}
where $\mathbf{E}^0$ denotes the initial embeddings that are fed to the first Transformer layer.

\subsubsection{Update Rules}
Based on the DAG representation of information flow \cite{abnar2020quantifying}, layer-wise aggregation is mathematically equivalent to matrix multiplication of intermediate maps. This approach allows for tracing token contributions over the entire model.
After initialization, we iteratively update the contribution map $\mathbf{C}^{l-1}$ using $\mathbf{U}^l$, which incorporates the effects of transformed tokens:
\begin{equation}
\mathbf{C}^l \leftarrow \mathbf{U}^l \cdot \mathbf{C}^{l-1} \quad \text{for} \quad l = 1, 2, \ldots, n_L.
\end{equation}
In this formula, $\mathbf{C}^{l-1}$ represents the map that has traced the token contributions up to the $(l-1)$-th layer but has yet to include the effects from the $l$-th layer, and $\mathbf{U}^{l}$ indicates the update map from the $l$-th layer.
Using this recursive formula, the map aggregates information about both token transformation and contextualization over the entire model. After applying the updates for all the layers, the final contribution map can be expressed as:
\begin{equation}
\mathbf{C}^{n_L} = \mathbf{U}^{n_L} \cdot \mathbf{U}^{n_L-1} \cdot \text{ } \cdots \text{ } \cdot \mathbf{U}^1 \cdot \mathbf{C}^0,
\end{equation}
% where $\mathbf{C}^0$ represents the initial state as described earlier.
This framework provides a cumulative understanding of how initial input tokens contribute to the final output, thereby faithfully localizing the most important pixels for Vision Transformers' predictions.

\subsection{Utilizing Contribution Map for Interpretation}
Vision Transformers often use designated tokens to perform specific tasks.
For example, ViT \cite{dosovitskiy2020image} performs image classification based on a $[CLS]$ token.
This special token becomes significant in the final prediction as it integrates information from all the inputs.
To interpret the model's prediction, one can analyze the contribution of each initial input token to the final $[CLS]$ token \cite{jain2019attention, abnar2020quantifying, chefer2021transformer, qiang2022attcat}.
In our proposed TokenTM, this is embodied in the overall contribution map $\mathbf{C}^{n_L}$.
Specifically, the row in $\mathbf{C}^{n_L}$ that corresponds to the $[CLS]$ token represents the influences of original tokens.
As illustrated in Figure \ref{figure3}, we extract this row and reshape it to the image's spatial dimensions, forming a contribution heatmap.
This heatmap highlights regions that are highly influential in determining the prediction, serving as a post-hoc explanation of the model.

%% file: sec/5_experiment.tex
\section{Experiments}
% In this section, we first introduce experimental setups. Then we demonstrate the superiority of our TokenTM qualitatively and quantitatively. Lastly, we take an ablation study on the proposed method.

\subsection{Baseline Methods}
We consider baseline methods that are widely used and applicable from three types.
\textbf{(i)}
Gradient-based: Grad-CAM \cite{selvaraju2017grad},
\textbf{(ii)}
Attribution-based: LRP \cite{binder2016layer}, Conservative LRP \cite{ali2022xai}, and Transformer Attribution \cite{chefer2021transformer}), and
\textbf{(iii)}
Attention-based: Raw Attention \cite{jain2019attention}, Rollout \cite{abnar2020quantifying}, ATTCAT \cite{qiang2022attcat}, and GAE \cite{chefer2021generic}).
% Details of experimental setups are provided in the supplementary.

\begin{table}[t]
\small
\centering
    \resizebox{0.85\columnwidth}{!}{%
    \begin{tabular}{ccccc}
        \toprule
        &Pix. Acc. $\uparrow$ & mAP $\uparrow$ & mIoU $\uparrow$ \\
        \midrule
        Grad-CAM \cite{selvaraju2017grad} &67.37 &79.20 &46.13 \\

        LRP \cite{binder2016layer} &50.96 &55.87 &32.82 \\

        Cons. LRP \cite{ali2022xai} &64.06 &68.01 &36.23 \\

        Trans. Attr. \cite{chefer2021transformer} &79.17 &85.81	&61.02 \\

        Raw Att. \cite{jain2019attention} &59.07 &76.51 &36.36 \\

        Rollout \cite{abnar2020quantifying} &59.01 &73.76 &39.43 \\

        ATTCAT \cite{qiang2022attcat} &43.88 &48.38 &27.90 \\

        GAE \cite{chefer2021generic} &79.05	&85.71 &61.56 \\

        {\textbf{Ours}} &\textbf{81.79}	&\textbf{86.45}	&\textbf{64.22} \\
        \bottomrule
\end{tabular}
}
\caption{Results of Localization Ability on ImageNet-Seg. \cite{guillaumin2014imagenet}.}
\vspace{-0.4cm}
\label{table1}
\end{table}

\subsection{Evaluated Properties}
% Following previous works \cite{chefer2021transformer, chefer2021generic, qiang2022attcat}, we evaluate three properties of the explanation methods.

\textbf{Localization Ability.}
This property measures how well an explanation can localize the foreground object recognized by the model.
Intuitively, a reliable explanation should be object-centric, \emph{i.e.}, accurately highlighting the object that the model uses to make the decision.
Following previous works \cite{selvaraju2017grad, barkan2021grad, chefer2021transformer, chefer2021generic}, we perform segmentation on the ImageNet-Segmentation dataset \cite{guillaumin2014imagenet}, using DeiT \cite{touvron2021training}, a prevalent Vision Transformer model.
We regard the explanation heatmaps as preliminary semantic signals.
Then we use the average value of each heatmap as a threshold to produce binary segmentation maps.
Based on the ground-truth maps, we evaluate the performance on three metrics: Pixel-wise Accuracy, mean Intersection over Union (mIoU), and mean Average Precision (mAP).
% Note that (c) is computed on the preliminary signals, thus can eliminate the effect of different threshold methods.

\noindent\textbf{Impact on Accuracy.}
This aspect focuses on how well an explanation captures the correlations between pixels and the model's accuracy.
We assess this property by perturbation tests on CIFAR-10, CIFAR-100 \cite{krizhevsky2009learning}, and ImageNet \cite{russakovsky2015imagenet}.
The pre-trained DeiT model is used to perform both positive and negative perturbation tests \textit{w.r.t.} the predicted and the ground-truth class.
Specifically, these tests include two stages.
First, we produce explanation maps for the class we want to visualize.
Second, we gradually remove pixels and evaluate the resulting mean top-1 accuracy.
In positive tests, pixels are removed from the most important to the least, while in negative tests, the removal order is reversed.
A faithful explanation should assign higher scores to pixels that contribute more.
Thus, a severe drop in the model's accuracy is expected in positive tests, while we expect the accuracy to be maintained in negative tests.
For quantitative evaluation, we compute the Area Under the Curve (AUC) \cite{atanasova2020diagnostic} \textit{w.r.t.} the accuracy curve of different perturbation levels.

\noindent\textbf{Impact on Probability.}
This further measures how well the explanations capture important pixels for the model's predicted probabilities.
Similarly, it is evaluated by perturbation tests.
We report results on Area Over the Perturbation Curve (AOPC) and Log-odds score (LOdds) \cite{shrikumar2017learning, nguyen2018comparing, chen2020generating}.
These metrics quantify the average change of output probabilities \textit{w.r.t.} the predicted label.
For comprehensive evaluations, we report results on ViT-B and ViT-L \cite{dosovitskiy2020image}.

\begin{table}[t]
\small
\centering
    \resizebox{0.9\columnwidth}{!}{%
    \begin{tabular}{cccccc}
        \toprule
        % &\multicolumn{4}{c}{Impact on Accuracy}\\
        % \cmidrule(lr){2-5}
        &\multicolumn{2}{c}{Predicted label} &\multicolumn{2}{c}{GT label}\\
        \cmidrule(lr){2-3}\cmidrule(lr){4-5}
        &Neg. $\uparrow$ & Pos. $\downarrow$ & Neg. $\uparrow$ & Pos. $\downarrow$ \\
        \midrule
        Grad-CAM \cite{selvaraju2017grad} &61.24 &47.32 &61.26 &47.34\\

        LRP \cite{binder2016layer} &72.62 &72.14 &72.54 &72.13\\

        Cons. LRP \cite{ali2022xai} &44.92 &49.15 &45.76 &47.87\\

        Trans. Attr. \cite{chefer2021transformer} &67.65 &43.64 &65.82 &43.17\\

        Raw Att. \cite{jain2019attention} &58.46 &47.17 &58.46 &47.17\\

        Rollout \cite{abnar2020quantifying} &64.88 &42.89 &64.88 &42.89\\

        ATTCAT \cite{qiang2022attcat} &41.78 &42.66 &41.91 &42.52 \\

        GAE \cite{chefer2021generic} &71.67 &37.77 &71.81 &37.70 \\

        {\textbf{Ours}} &\textbf{74.90} &\textbf{37.14} &\textbf{75.05} &\textbf{36.48} \\
        \bottomrule
\end{tabular}
}
\caption{Results of Impact on Accuracy on CIFAR-10 \cite{krizhevsky2009learning}.}
\vspace{-0.4cm}
\label{table2}
\end{table}

\begin{table}[t]
\small
\centering
    \resizebox{0.9\columnwidth}{!}{%
    \begin{tabular}{cccccc}
        \toprule
        % &\multicolumn{4}{c}{Impact on Accuracy}\\
        % \cmidrule(lr){2-5}
        &\multicolumn{2}{c}{Predicted label} &\multicolumn{2}{c}{GT label}\\
        \cmidrule(lr){2-3}\cmidrule(lr){4-5}
        &Neg. $\uparrow$ & Pos. $\downarrow$ & Neg. $\uparrow$ & Pos. $\downarrow$ \\
        \midrule
        Grad-CAM \cite{selvaraju2017grad} &47.18 &41.91 &47.28 &41.78\\

        LRP \cite{binder2016layer} &46.76 &46.31 &46.80 &46.29\\

        Cons. LRP \cite{ali2022xai} &27.11 &27.99 &26.84 &27.42\\

        Trans. Attr. \cite{chefer2021transformer} &44.65 &24.99 &44.19 &23.90\\

        Raw Att. \cite{jain2019attention} &37.23 &28.62 &37.23 &28.62\\

        Rollout \cite{abnar2020quantifying} &41.34 &24.96 &41.34 &24.96\\

        ATTCAT \cite{qiang2022attcat} &23.52 &22.15 &23.90 &21.73 \\

        GAE \cite{chefer2021generic} &52.14 &17.52 &52.65 &17.32 \\

        {\textbf{Ours}} &\textbf{54.78} &\textbf{16.07} &\textbf{55.30} &\textbf{15.89} \\
        \bottomrule
\end{tabular}
}
\caption{Results of Impact on Accuracy on CIFAR-100 \cite{krizhevsky2009learning}.}
\vspace{-0.4cm}
\label{table3}
\end{table}

\begin{table}[t]
\small
\centering
    \resizebox{0.9\columnwidth}{!}{%
    \begin{tabular}{cccccc}
        \toprule
        % &\multicolumn{4}{c}{Impact on Accuracy}\\
        % \cmidrule(lr){2-5}
        &\multicolumn{2}{c}{Predicted label} &\multicolumn{2}{c}{GT label}\\
        \cmidrule(lr){2-3}\cmidrule(lr){4-5}
        &Neg. $\uparrow$ & Pos. $\downarrow$ & Neg. $\uparrow$ & Pos. $\downarrow$ \\
        \midrule
        Grad-CAM \cite{selvaraju2017grad} &43.84 &26.11 &44.06 &25.87\\

        LRP \cite{binder2016layer} &42.55 &41.29 &42.56 &41.29\\

        Cons. LRP \cite{ali2022xai} &33.24 &34.35 &34.33 &34.03\\

        Trans. Attr. \cite{chefer2021transformer} &57.48 &19.00 &58.26 &18.38\\

        Raw Att. \cite{jain2019attention} &48.70 &25.44 &48.70 &25.44\\

        Rollout \cite{abnar2020quantifying} &43.23 &32.34 &43.23 &32.34\\

        ATTCAT \cite{qiang2022attcat} &32.34 &31.71 &33.59 &30.68 \\

        GAE \cite{chefer2021generic} &57.53 &15.88 &58.85 &15.20 \\

        {\textbf{Ours}} &\textbf{58.12} &\textbf{15.75} &\textbf{59.36} &\textbf{15.09} \\
        \bottomrule
\end{tabular}
}
\caption{Results of Impact on Accuracy on ImageNet \cite{russakovsky2015imagenet}.}
\vspace{-0.4cm}
\label{table4}
\end{table}

\subsection{Experimental Results}
\textbf{Qualitative Evaluation.}
As shown in Figure \ref{figure4}, interpretation heatmaps of TokenTM are more precise and exhaustive.
Rather than indicating plain background areas as rationale, the integration of cross-layer token transformation information in TokenTM effectively eliminates noisy regions, allowing a more object-focused analysis.
More visualizations are provided in the supplementary.

\noindent\textbf{Quantitative Evaluation.}
\textbf{(i)}
\textbf{Localization Ability.}
Table \ref{table1} reports the segmentation results on the ImageNet-Segmentation dataset. Our proposed TokenTM significantly outperforms all the baselines on pixel accuracy, mIoU, and mAP, demonstrating its stronger localization ability.
\textbf{(ii)}
\textbf{Impact on Accuracy.}
Tables \ref{table2}, \ref{table3}, and \ref{table4}
show the results of the perturbation tests for both classes.
For positive tests, a lower AUC indicates superior performance, while for negative tests, a higher AUC is desirable. The results underscore the superiority of our method.
% Also, our explanation result is class-specific, which is verified by the gain in performance when the target label is used (compared to the predicted label).
\textbf{(iii)}
\textbf{Impact on Probability.}
Table \ref{table5} reports the results of perturbation tests assessing the model's output probability.
Our TokenTM achieves the best performance on Vision Transformers variants.

\begin{table}[t]
\centering
    \resizebox{0.9\columnwidth}{!}{%
    \Large
    \begin{tabular}{ccccccc}
        \toprule
        % &\multicolumn{6}{c}{Impact on Probability}\\
        % \cmidrule(lr){2-7}
        &\multicolumn{3}{c}{AOPC $\uparrow$} &\multicolumn{3}{c}{LOdds $\downarrow$}\\
        \cmidrule(lr){2-4}\cmidrule(lr){5-7}
        & ViT-B \cite{dosovitskiy2020image} & ViT-L \cite{dosovitskiy2020image} & ViT-B \cite{dosovitskiy2020image} & ViT-L \cite{dosovitskiy2020image} \\
        \midrule
        Grad-CAM \cite{selvaraju2017grad} & \textcolor{black}{0.557} & \textcolor{black}{0.457} & \textcolor{black}{-4.020} & \textcolor{black}{-3.023}\\

        LRP \cite{binder2016layer} & \textcolor{black}{0.474} & \textcolor{black}{0.508} & \textcolor{black}{-3.259} & \textcolor{black}{-3.634}\\

        Cons. LRP \cite{ali2022xai} & \textcolor{black}{0.613} & \textcolor{black}{0.614} & \textcolor{black}{-4.272} &\textcolor{black}{-4.443}\\

        Trans. Attr. \cite{chefer2021transformer} & \textcolor{black}{0.721} & \textcolor{black}{0.716} & \textcolor{black}{-5.622} & \textcolor{black}{-5.541}\\

        Raw Att. \cite{jain2019attention} & \textcolor{black}{0.652} & \textcolor{black}{0.655} & \textcolor{black}{-4.911} & \textcolor{black}{-4.961}\\

        Rollout \cite{abnar2020quantifying} & \textcolor{black}{0.688} & \textcolor{black}{0.689} & \textcolor{black}{-5.259} & \textcolor{black}{-5.183}\\

        ATTCAT \cite{qiang2022attcat} & \textcolor{black}{0.641} & \textcolor{black}{0.645} & \textcolor{black}{-4.461} & \textcolor{black}{-4.622}\\

        GAE \cite{chefer2021generic} & \textcolor{black}{0.714} & \textcolor{black}{0.699} & \textcolor{black}{-5.437} & \textcolor{black}{-5.601}\\

        % {\textbf{Oursl2}} &\underline{\textbf{\textcolor{black}{0.297}}} / \underline{\textbf{\textcolor{blue}{0.651}}} / \underline{\textbf{\textcolor{red}{0.822}}} &\underline{\textbf{\textcolor{black}{0.725}}} / \underline{\textbf{\textcolor{blue}{0.737}}} / \underline{\textbf{\textcolor{red}{0.861}}} &\underline{\textbf{\textcolor{black}{0.724}}} / \underline{\textbf{\textcolor{blue}{0.719}}} / \underline{\textbf{\textcolor{red}{0.832}}} &\underline{\textbf{\textcolor{black}{-4.514}}} / \underline{\textbf{\textcolor{blue}{-3.466}}} / \underline{\textbf{\textcolor{red}{-5.268}}} &\underline{\textbf{\textcolor{black}{-5.667}}} / \underline{\textbf{\textcolor{blue}{-3.605}}} / \underline{\textbf{\textcolor{red}{-5.486}}} &\underline{\textbf{\textcolor{black}{-5.705}}} / \underline{\textbf{\textcolor{blue}{-3.548}}} / \underline{\textbf{\textcolor{red}{-4.260}}}\\

        \textbf{Ours} &\textbf{0.764} &\textbf{0.726} &\textbf{-5.781} &\textbf{-5.755}\\
        \bottomrule
\end{tabular}
}
\caption{Results of Impact on Probability on more Vision Transformer models.}
\vspace{-0.2cm}
\label{table5}
\end{table}

\begin{table}
    \centering
    \resizebox{1\columnwidth}{!}{%
    \Large
    \begin{tabular}{cccccc}
        \toprule
        &\multicolumn{3}{c}{Segmentation} &\multicolumn{2}{c}{Perturbation Test}\\
        \cmidrule(lr){2-4}\cmidrule(lr){5-6}
        &Pix. Acc. $\uparrow$ & mAP $\uparrow$ & mIoU $\uparrow$ & Neg. $\uparrow$ & Pos. $\downarrow$\\
        \midrule
        Baseline &52.94	&67.78	&33.42 &44.92 &31.95\\
        + AF &76.30	&85.28	&58.34	&55.25 &16.98\\
        + AF + L &77.50	&85.39	&59.59	&55.03 &16.28\\
        + AF + L + NECC &\textbf{78.22}	&\textbf{85.69}	&\textbf{60.29}	&\textbf{55.42} &\textbf{16.18}\\
        \bottomrule
    \end{tabular}
    }
\caption{Results of ablation study on the proposed components, \emph{i.e.}, transformation measurement (L and NECC) and aggregation framework (AF).}
\vspace{-0.2cm}
\label{table6}
\end{table}

\subsection{Ablation Study}
\noindent \textbf{Ablation Study on Proposed Components.}
We conduct ablation studies on the effect of our proposed transformation measurement (L and NECC) and aggregation framework (AF).
We perform experiments on ImageNet \cite{russakovsky2015imagenet}, based on ViT-B \cite{dosovitskiy2020image}.
Four variants are considered in our ablative experiment:
\textbf{(i)}
Baseline, which merely applies Eq. \eqref{eq8} to the input tokens without applying our proposed components,
\textbf{(ii)}
Baseline + AF, which initializes the contribution map as an identity matrix and aggregates across multiple layers using $\mathbf{U}^l$, where our proposed relative lengths and NECC measurements are discarded,
\textbf{(iii)}
Baseline + AF + L, which additionally use relative length measurement in $\mathbf{U}^l$ while still excluding the NECC terms, and
\textbf{(iv)}
Baseline + AF + L + NECC, which is our TokenTM.
As shown in Table \ref{table6}, each proposed component improves the performance on both segmentation and perturbation tests, validating their effectiveness in the Vision Transformer explanation.

\begin{figure}[t]
  \centering
  \includegraphics[width=0.9\linewidth]{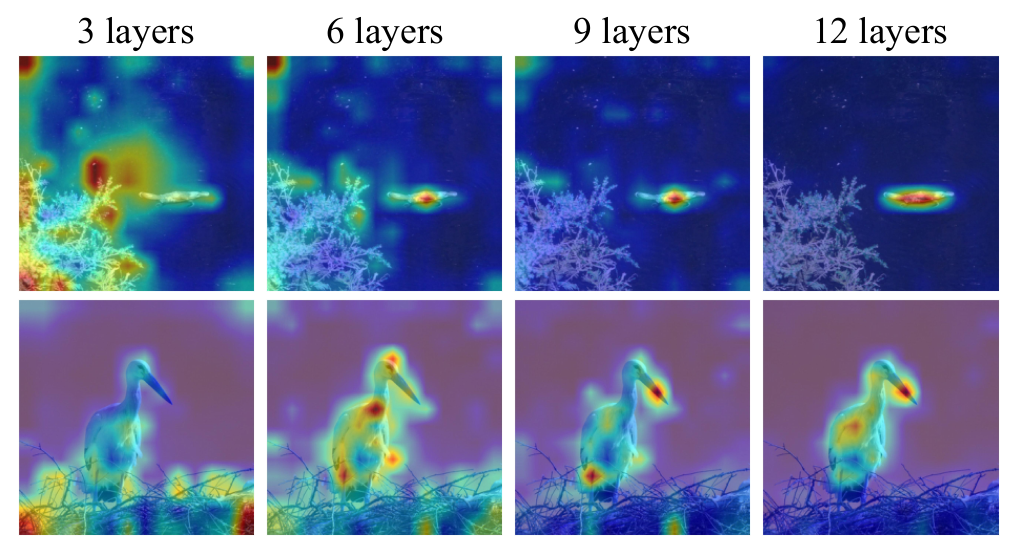}
  \caption{Visualizations of the localization process using our TokenTM. As more layers are incorporated into the aggregation, the heatmaps become increasingly focused, particularly in the regions surrounding the object recognized by the model.}
  \vspace{-4mm}
  \label{figure5}
\end{figure}

\noindent \textbf{Ablation Study on Aggregation Depth.}
To better understand the cumulative nature of Vision Transformers using TokenTM, we study the impact of varying the aggregation depth within our method.
Specifically, we increase the number of aggregated layers from the initial few to the entire network depth.
Table \ref{table7} reports the ablative results on ImageNet \cite{russakovsky2015imagenet}, using ViT-B \cite{dosovitskiy2020image}, where `n layers' denotes a variant of our TokenTM that (only) aggregates the first n layers of the model.
The results reveal a consistent enhancement in performance with increased layer aggregation depth, which suggests that a deeper aggregation of token transformation and contextualization effects is pivotal for capturing the true rationale of the model's inference.
Furthermore, Figure \ref{figure5} illustrates the reasoning procedure of the Vision Transformer.
As the aggregation encompasses more layers, the heatmaps progressively refine the focused areas and become more concentrated on the object, localizing the rationale behind the model's prediction.

\begin{table}
    \centering
    \resizebox{1\columnwidth}{!}{%
    \Large
    \begin{tabular}{cccccc}
        \toprule
        &\multicolumn{3}{c}{Segmentation} &\multicolumn{2}{c}{Perturbation Test}\\
        \cmidrule(lr){2-4}\cmidrule(lr){5-6}
        &Pix. Acc. $\uparrow$ & mAP $\uparrow$ & mIoU $\uparrow$ & Neg. $\uparrow$ & Pos. $\downarrow$\\
        \midrule
        3 layers &63.63	&76.40	&42.99 &43.32 &22.57\\
        6 layers &68.38	&80.35	&48.66	&48.28 &20.17\\
        9 layers &70.46	&81.74	&51.11	&50.42 &19.24\\
        12 layers &\textbf{78.22}	&\textbf{85.69}	&\textbf{60.29}	&\textbf{55.42} &\textbf{16.18}\\
        \bottomrule
    \end{tabular}
    }
\caption{Results of ablation study on the aggregation depth.}
\vspace{-0.2cm}
\label{table7}
\end{table}

%% file: sec/6_conclusion.tex
\section{Conclusions}
In this paper, we explored the challenge of explaining Vision Transformers.
Upon reinterpretation, we expressed Transformer layers using a generic form in terms of weighted linear combinations of original and transformed tokens.
This new perspective revealed a principal issue in existing post-hoc methods:
they solely rely on attention weights and ignore significant information from token transformations, which leads to misunderstandings of rationales behind the models' decisions.
To tackle this issue, we proposed TokenTM, an explanation method comprising a token transformation measurement and an aggregation framework.
Quantifying the contributions of input tokens throughout the entire model, our method generates more faithful post-hoc interpretations for Vision Transformers.
% Experiments on segmentation and perturbation tests across various datasets and Vision Transformer models demonstrated the superiority of our TokenTM compared to state-of-the-art explanation methods.

\noindent \textbf{Acknowledgments:} This research is supported by NSF IIS-2309073 and ECCS-212352101. This article solely reflects the opinions and conclusions of its authors and not the funding agents.